**Mapping Human-Agent Co-Learning and Co-Adaptation: A Scoping Review**

Shruti Kumar, Xiaoyu Chen, Xiaomei Wang*

# *Introduction*

## Background

Robots must adapt to human team behaviors to establish successful collaboration patterns as they become more prevalent. Many adaptations manifest through subtle and unconscious interactions, which are challenging to observe [1]. In service environments like restaurants, robots like the WAITER robot must adapt to human behaviors for effective collaboration. It requires flexible mappings between human instructions and robot actions, enabling intuitive interactions [2], [3], [4], [5]. Mutual adaptation, where both the robot and human adjust their behaviors [6], is crucial for efficient task completion and is guided by computational models like the Bounded-Memory Adaptation Model [7]. This co-evolution process shapes the tasks and environments, enhancing human learning and adaptation [3], [4], [5], [8], [9], [10], [11], [12], [13], [14], [15], [16], [17], [18], [19], [20], [21], [22], [23], [24], [25], [26], [27], [28], [29], [30], [31], [32]. In BCI systems, mutual adaptation enhances collaboration by creating a co-adaptation loop where the human and the system learn from each other. Techniques like ErrPs-based mediation improve the system's response by adjusting actions based on real-time EEG feedback [33], [34]. The FML-based Reinforcement Learning Agent (FRL-Agent) further facilitates continuous learning and adaptation, crucial in applications like neural prosthetics and rehabilitation devices [1], [6], [13], [35], [36], [37], [38], [39], [40], [41], [42], [43], [44], [45], [46], [47], [48], [49], [50], [51], [52]. Robots are increasingly used as learning partners, where mutual adaptation plays a significant role. For example, robots can co-learn with humans in various tasks, from motor tasks to language acquisition, using models that allow humans and robots to adjust their behaviors based on the interaction [53], [54], [55]. These interactions enhance the learning experience and promote the development of trust and effective strategies in human-robot teams [56], [57], [58], [59]. In manufacturing and assistive technology, mutual adaptation ensures seamless human-robot collaboration. Systems like the Adaptive Fuzzy Neural Agent (AFNA) use co-learning mechanisms to optimize performance in intelligent factories and rehabilitation [35], [60]. These systems rely on real-time data sharing and adaptive control algorithms to respond to human inputs, ensuring safe and efficient operations [31], [61], [62], [63]. Shared autonomy integrates human input with robot autonomy to improve task performance. Mutual adaptation is critical in these scenarios, as humans and robots must adjust their actions based on feedback and shared goals [64], [65]. This approach is efficient in applications like assistive robotics, where the robot helps humans achieve better outcomes by guiding them toward optimal strategies [36], [66], [67]. Mutual learning between humans and AI systems supports collaborative decision-making in various domains, such as healthcare and education. AI systems adapt to human expertise and preferences, helping teams to co-evolve and improve over time [68], [69], [70]. The use of co-learning frameworks enhances the ability of AI to assist in complex decision-making processes, making these systems more responsive and effective [71], [72], [73], [74]. Interactive systems incorporating mutual adaptation, such as teleoperation and assistive devices, improve user experience by learning from user behavior and adjusting system responses accordingly [75], [76]. This approach is essential for developing intuitive interfaces that effectively collaborate with users in tasks requiring fine motor control or complex decision-making [63], [77], [78]. Mutual adaptation fosters engaging and effective interactions between humans and robots in social play and edutainment. By learning from human behavior, robots can participate in games, teach new skills, or serve as companions, enhancing the overall experience [53], [74], [79]. These interactions often involve co-learning processes that help both the human and the robot to improve their performance and adapt to new challenges [53], [58], [80].

### Objective

Several papers have delved into the challenges of human-AI-robot co-learning and co-adaptation [1], [38], [39], [81]. It has been noted that the terminology used to describe this collaborative relationship in existing studies needs to be more consistent [39]. For example, the prefix "co" is used interchangeably to represent both "collaborative" and "mutual," and the terms "co-learning" and "co-adaptation" are sometimes used interchangeably. However, they can reflect subtle differences in the focus of the studies.

The current scoping review's primary research question (RQ1) aims to gather existing papers discussing this collaboration pattern and examine the terms researchers use to describe this human-agent relationship.

Given the relative newness of this area of study, we are also keen on exploring the specific types of intelligent agents and task domains that have been considered in existing research (RQ2). This exploration is significant as it can shed light on the diversity of human-agent interactions, from one-time to continuous learning/adaptation scenarios. It can also help us understand the dynamics of human-agent interactions in different task domains, guiding our expectations towards research situated in dynamic, complex domains.

Our third objective (RQ3) is to investigate the cognitive theories and frameworks that have been utilized in existing studies to measure human-agent co-learning and co-adaptation. This investigation is crucial as it can help us understand the theoretical underpinnings of human-agent collaboration and adaptation, and it can also guide us in identifying any new frameworks proposed specifically for this type of relationship.

## *Methods*

We conducted a scoping review to explore the current literature on Human and intelligent agents working in a collaborative space as a hybrid team to "co-learn" (also known as "co-adapt"), which is relatively new and intriguing to the researchers. Scoping reviews are an effective and valuable strategy for synthesizing emerging concepts and topics in a specific domain. Our review adheres to the PRISMA-ScR (Preferred Reporting Items for Systematic Reviews and Meta-Analyses Extension for Scoping Reviews) guidelines (Multimedia Appendix 1). We meticulously searched all peer-reviewed publications in the Web of Science, Engineering Village, and EBSCOhost database published before January 2024 to identify studies meeting the eligibility criteria within the scope of our review.

**Search Strategy**

Our search strategy was not just systematic, but also comprehensive. We meticulously crafted search terms to ensure we included all relevant papers in the database. These keywords were selected after a thorough preliminary literature review and were further refined based on feedback from content experts and our institution's librarian. Our collaboration with the librarian focused on refining the search strategy and ensuring we considered all papers related to human-agent/AI-based co-learning and mutual adaptation in our review. We used a hierarchical structure similar to a tree to optimize our search, starting from general (branch) terms and moving towards more specific (leaf) ones. We included synonyms in our search strategy to ensure a comprehensive search. For example, "Intelligent Agents," which is a branch term, was also represented by synonyms such as "Artificial Intelligence," "Robot," "Machine," and so on. We also accounted for leaf terms like "Machine Learning" and "Reinforcement Learning" as part of our search strategy. Figure 1 illustrates the different combinations of these terms and their variations, along with the Boolean (AND and OR) operators used to identify all relevant studies that met our criteria. For example, "Human-Intelligent Agents {OR} Human-Artificial Agents {OR} ..." {AND} "Co-learning {OR} Co-adaptation {OR} ..." were used to refine our search.

## Definition for Human- Intelligent Agent Co-learning/Co-adaptation

The use of increasingly advanced AI technology is reshaping how individuals and teams learn and carry out their tasks. In hybrid teams, people work alongside artificially intelligent partners [38]. To leverage the unique strengths and weaknesses of a human and an intelligent agent, a hybrid team should be constructed based on principles that facilitate successful human-machine learning, interaction, and collaboration based on the principles of co-learning and mutual adaptation, where both the members of the team are benefited by learning from each other.

Zoelen [39], describes co-learning as developing a cohesive team that necessitates ongoing collaborative learning among all team members. This process encompasses two alternating iterative stages: first, partners adapt their behavior to the task and each other (co-adaptation); second, partners sustain effective behavior through communication. For example, in his study, he recognized the recurrent behaviors indicative of co-learning by a task context conducive to the emergence of behavioral adaptation from interactions between humans and robots.

Similarly, Nikolaidis[2] defines mutual adaptation between an agent and a human as ability of members of the team to co-exist in the same physical space with the aim to make more informed decisions and become trustworthy team members to achieve improved performances for the given task.

In this paper, our focus is on reviewing human-agent collaboration which is a two-way sharing of responsibility and processing of information which determines a conducive environment for co-learning and mutual adaption. Furthermore, this review reflects on the domains and the important cognitive frameworks involved for co-adaptation during the human-agent teaming.

**Figure 1.** Terms used for the Conceptual framework of the scoping review.

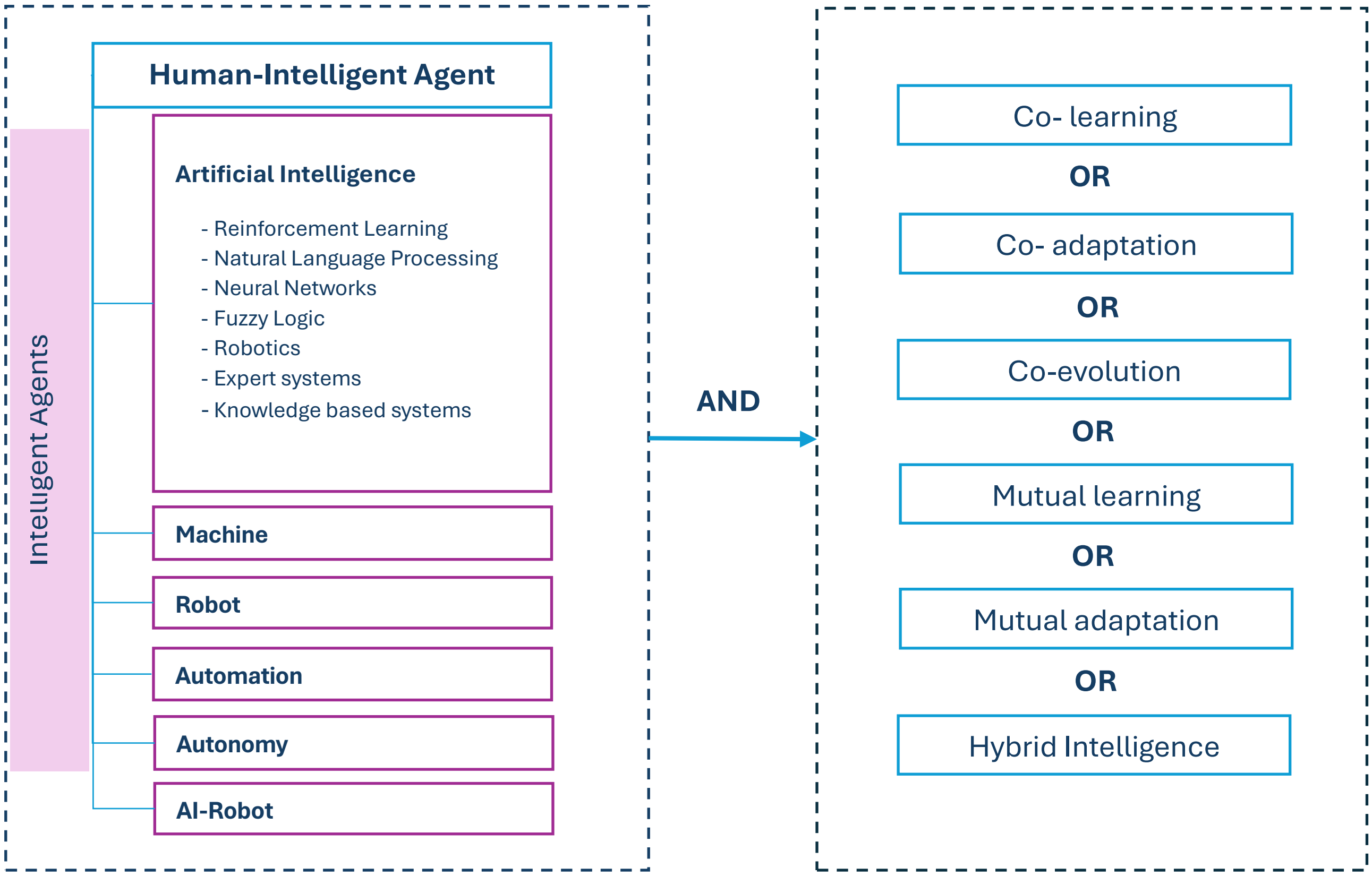

## Inclusion and Exclusion Criteria

This review included peer-reviewed studies that met the following primary conditions:

1. The study focused on a Human-Intelligent Agent-based co-learning and mutual adaptation that involved two-way learning, adaptation, and evolution, which were the highlights of this study.

We excluded studies with the following criteria:

1. Studies related to co-creation (typically using generative AI for co-creating art products).
2. Those focusing on motor adaptation (e.g., mimetic robot).
3. Studies centered on computational linguistics.
4. Those discussing the relationship between human society and AI instead of specific Humans and agents. (co-evolution)
5. For comparison, the study included those that state "co-adaptation" and "mutual adaptation" but describe only the agent's adaptation to Humans, not humans and agents adapting to each other, for example, education/training purposes.

6. Studies not published in English.

### Study data collection and synthesis

We began the paper selection process by evaluating the eligibility of the studies based on the authors' inclusion and exclusion criteria. First, we removed duplicate papers and screened the remaining studies by reviewing their titles and abstracts. Next, we used an Excel spreadsheet to take detailed notes on each study. This spreadsheet contained information about the study's objective, application based on the domain and adaptation style, cognitive theories and frameworks, type of intelligent agent, the mutual performance of the Human and the agent, sample size, and additional notes. Any discrepancies were resolved through discussion to ensure fairness, requiring consensus between the reviewers. We standardized information from each paper after collecting the desired information using a PRISMA-data abstraction form. We used this method to finalize the selection of studies for review. Figure 2 displays the details regarding the study selection process.

## Results

### Study Selection

Two authors meticulously compiled the papers selected for this scoping review in a systematic flowchart representation, as shown in Figure 2. Using specific keywords, the initial search on Web of Science, Engineering Village, and EBSCOhost resulted in 373 papers. We organized them using the Mendeley Reference Manager (v2.118.0), thus helping us filter and remove duplicate documents (131/373 = 35.12%). Further screening based on titles and abstracts left the authors with 92 papers for a full-text review. The last round of screening was based on inclusion and exclusion criteria, as mentioned in the respective section, resulting in the selection of 92 papers (38.01% of the original 242) for full-text review. Finally, based on the eligibility, the authors removed 10.86% (10/92) of papers due to the unavailability and based on the full-text review. Additionally, later in the review, we include a discussion on applications of co-adaptation, covering two-way adaptation (84.14%), one-way adaptation (6.09%), both two-way and one-way (4.87%), and cases where information was not available (4.87%).

**Figure 2.** PRISMA flowchart for systematic searching and selecting the papers.

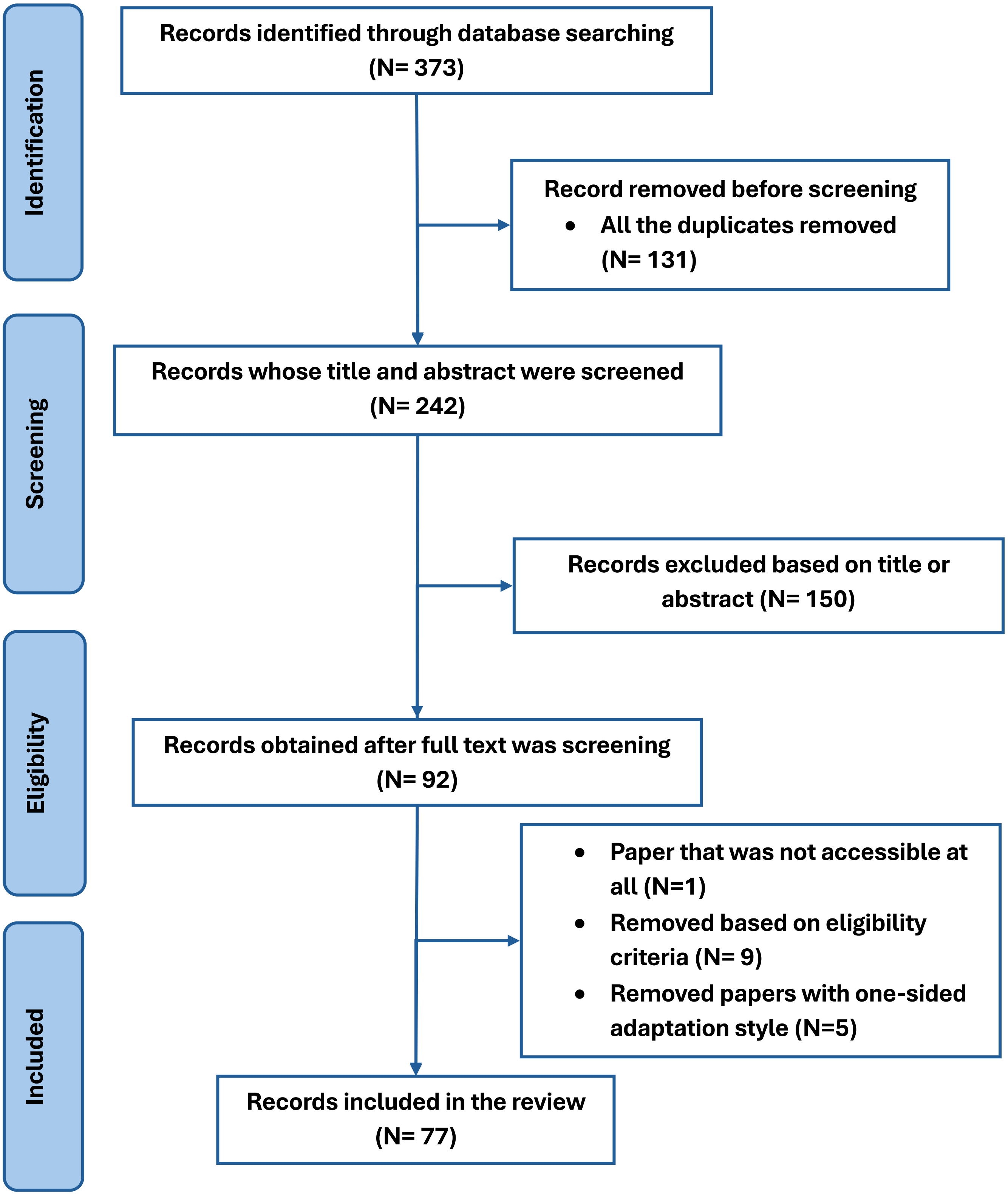

Table 1 provides an overview of 77 selected reports detailing the study's objective, the application domain and style of the human-intelligent agent system, and the cognitive theories and frameworks used by the hybrid team. Most of the work, over 50%, has been conducted in the last four years, indicating a need for future expansion in this emerging field of study. Notably, 54.54% (42/77) of the papers were published within the last 2 to 3 years, demonstrating significant and recent growth in this research area.

**Table 1.** The study design and reported findings of final selected papers.

| **Study, year, place of Publication** | **Objective** | **Application** | | **Cognitive theories and frameworks used** | **Sample size, n** |
|---|---|---|---|---|---|
| | | **Domain/Tasks** | **Adaptation style** | | |
| Emma M. et al [1], 2021, Netherlands | To explore and identify interaction patterns of tangible co-adaptations in human-robot team behaviors to enhance collaborative performance. | Search and navigation, interaction patterns | Two-way | • Mental model<br>• Decision making<br>• Trust | 18 |
| Stefanos Nikolaidis et al [2], 2017, United States | To develop a computational framework for mutual adaptation between humans and robots in collaborative tasks, enhancing team effectiveness and trust | Bounded-Memory Adaptation Model | One-way and Two-way both | • Decision making<br>• Trust | 170 |
| Xu Yong et al [3], 2009, Japan | To introduce and evaluate a learning model for mutual adaptation in human-robot collaboration. | Restaurant-WAITER (Waiter Agent Interactive Training Experimental Restaurant) | Two-way | • Decision making<br>• Trust<br>• Mental model | 10 |
| Xu Yong et al [4], 2011, Japan | To investigate the principal features of active adaptation in human-agent collaboration using a developed experimental environment. | Virtual Restaurants-WAITER (Waiter Agent Interactive Training Experimental Restaurant) | Two-way | • Decision making<br>• Trust<br>• Mental model | 37 |
| Xu Yong et al [5], 2009, Japan | To disclose the essence of mutual adaptation in | WAITER (Waiter Agent | Two-way | • Decision making | 25 |

| Study, year, place of Publication | Objective | Application | | Cognitive theories and frameworks | Sample size, n |
|---|---|---|---|---|---|
| | human-agent interaction by designing a waiter agent task and conducting experiments. | Interactive Training Experimental Restaurant) | | • Trust<br>• Mental model | |
| Yong Xu et al [6], 2012, Japan | To elucidate the conditions required to induce mutual adaptation in human-agent collaborative interactions using the “WAITER” experimental environment. | WAITER (Waiter Agent Interactive Training Experimental Restaurant) | Two-way | • Decision making<br>• Trust<br>• Mental model | 37 |
| Ramya Ramakrishnan et al [7], 2017, United States | To design and evaluate a computational learning model that enables human-robot teams to co-develop joint strategies for performing novel tasks through perturbation training. | Adaptive perturbation training, team-training strategy | Two-way | • Decision making<br>• Trust<br>• Mental model<br>• Team performance | 48 |
| Döppner et al [8], 2019, Germany | To investigate the evolution of human-machine collaboration (HMC) and explore patterns of symbiotic co-evolution in a real-world business context | Air cargo logistics - Artifact for human-machine collaboration (HMC) | Two-way | • Decision making | NA |
| Burke C. S. et al [9], 2006, United States | It provides a multidisciplinary, multilevel, and multiphasic conceptualization of team adaptation, defining adaptive team performance and its emergent nature from a theoretical standpoint. | NA | NA | • Teamwork<br>• Team effectiveness<br>• Shared mental model<br>• Safety | NA |
| Hendrik Buschmeier and Stefan Kopp [10], 2013, Germany | To present a computational model that enables an artificial conversational agent to estimate and adapt to the user's mental state for better grounding in dialogue. | Symbol Grounding problem | Two-way | • Utterance | 1 |
| Guy Hoffman [11], 2019, United States | To codify and evaluate metrics for measuring human-robot fluency in shared-location teamwork. | Turn-taking system | NA | • Human–robot fluency metrics<br>• Task efficiency | 143 |

| Study, year, place of Publication | Objective | Application | | Cognitive theories and frameworks | Sample size, n |
|---|---|---|---|---|---|
| Lygerakis Fotios et al [12], 2021, Greece | To investigate the impact of different allocations of online and offline gradient updates on game performance and training time in a real-time collaborative setting with humans. | Marble-Maze game | Two-way | • Performance<br>• Total training time | 2 |
| Zhao Michelle et al [13], 2022, United States | To explore a framework for AI that adapts to individual human partners within a group, emphasizing the benefits of AI adaptation in collaborative settings. | Simulated Navigation task- search-and-rescue (SAR) environment | Two-way | • Performance<br>• Management<br>• Decision making<br>• Mental state<br>• Preference | 108 |
| Nikolaidis et al [14], 2016, United States, Singapore | To present a formalism for human-robot mutual adaptation in collaborative tasks to improve team effectiveness. | Table carrying task, Bounded memory Adaptation Model | Two-way | • Effectiveness<br>• Performance<br>• Trust | 69 |
| Youssef Khaoula et al [15], 2014 | To explore how social interaction evolves incrementally and can be formalized into a protocol of communication, using a dish robot in a Sociable Dining Table context that requires mutual adaptation. | Sociable Dining Table | One-way and Two-way both | • Adaptive behavior<br>• Communicatio n | 42 |
| Damian Dana D. et al [16], 2009 | To develop automated metrics for quantifying mutual adaptation in human-robot interaction within rehabilitation robotics to enhance collaboration efficiency. | Rehabilitation robotics | Two-way | • Efficiency<br>• Decision making<br>• Performance | 7 |
| Zhou Hua et al [17], 2021 | To develop a novel negotiation strategy using virtual fixtures to facilitate mutual adaptation between humans and robots in collaborative tasks where both parties have incomplete information. | Physical human-robot interactions, manipulation tasks, haptic negotiation | Two-way | • Cognitive load<br>• Performance<br>• Decision making | 36 |
| Ogata T et al [18], 2005 | To investigate mutual adaptation between | Navigation System, | Two-way | • Performance | 7 |

| Study, year, place of Publication | Objective | Application | | Cognitive theories and frameworks | Sample size, n |
|---|---|---|---|---|---|
| | humans and robots in an open-end interaction scenario using a specially developed humanoid robot, Robovie, focusing on incremental learning and navigation system dynamics. | dynamical system | | • Incremental learning<br>• Robustness,<br>• Operability<br>• Mental impressions | |
| De Santis Dalia [19], 2021 | To propose and explore a mathematical framework for studying co-adaptation between humans and adaptive interfaces in scenarios where the interface evolves based on user behavior, aiming to maximize interaction efficiency rather than optimizing specific task performance. | Body-machine interfaces | Two-way | • Task performance<br>• Interaction efficiency | 10 |
| Yamada S and Yamaguchi T [20], 2002 | To propose and investigate a framework for mutual mind reading between a user and a life-like agent, aiming to establish effective and natural communication by acquiring each other's mind mappings through a mutual mind reading game. | Mind reading game, Mind mapping | Two-way | • Effective and natural communication<br>• Emotional state<br>• Processing load<br>• Mind state<br>• Judgement | 8 |
| Madduri Maneeshika M. et al [21], 2022 | To propose and experimentally test a game-theoretic framework for optimizing the closed-loop performance of a myoelectric interface for continuous control, focusing on co-adaptation between the user and the decoder. | Neural interfaces- HMI, BCI, rehabilitation, 2D trajectory-tracking task, sEMG | Two-way | • Performance | 7 |
| Sciutti Alessandra and Sandini Giulio [22], 2017 | To understand how human motion features contribute to emergent coordination with another agent (the robot). While not explicitly stated, the investigation suggests a mutual | neuroscientific bases of social interaction | Two-way | • Efficiency | NA |

| Study, year, place of Publication | Objective | Application | | Cognitive theories and frameworks | Sample size, n |
|---|---|---|---|---|---|
| | adaptation scenario where the robot reacts contingently to human actions, influencing the interaction dynamics. | | | | |
| Gao Yuan [23], 2020 | To develop and understand efficient reinforcement learning algorithms for human-robot co-adaptation in interactive environments. | social robotics | Two-way | • Decision making<br>• Performance | NA |
| Young Carol et al [24], 2019 | To analyze and improve an expert-based online learning algorithm to avoid chatter when predicting human intention during human-robot interaction. | Online Co-Learning, expert based learning | Two-way | • Chatter<br>• Human intentions<br>• Internal state,<br>• Mind state | NA |
| O'Meara Sarah M. et al [25], 2021 | To investigate whether adding adaptive timing features and classifiers to an electromyography (EMG) interface can improve the performance of trained human subjects in cursor-control tasks. | myoelectric control | Two-way | • Performance<br>• Reliability | 48 |
| Warrier Rahul B. and Devasia Santosh [26], 2017 | To develop a data-based approach for improved iterative robot learning for output tracking from novice human-in-the-loop demonstrations. | Data-based iterative learning | Two-way | • Output tracking<br>• Robot learning<br>• Performance | 1 |
| Damiano Luisa and Dumouchel Paul [27], 2018 | To propose a theoretical perspective on anthropomorphism in social robotics and to develop a critical experimentally based ethical approach called "synthetic ethics. | Social robotics, Anthropology | NA | • Self-knowledge<br>• Moral growth | NA |
| Cress Ulrike and Kimmerle Joachim, [28], 2023 | To explore how Generative AI tools can be used for individual learning and collective knowledge construction through argumentative dialogs. | CSCL research | Two-way | • NA | NA |

| Study, year, place of Publication | Objective | Application | | Cognitive theories and frameworks | Sample size, n |
|---|---|---|---|---|---|
| Ahtinen Aino et al [29], 2023 | To explore and define a model for family-based collaborative learning (co-learning) about and with a social robot in home environments. | Social Robotics | Two-way | • Learning experience<br>• Task completion | 32 |
| Kitagawa Masahiro et al [30], 2016 | To investigate mutual adaptation between humans and robots through the imitation of human motions by robots, examining how this interaction affects human behavior. | human-robot interaction, imitation | One-way and Two-way both | • Sleep-time<br>• Motion-time | 18 |
| Mohammad Yasser and Nishida Toyoaki [32], 2008 | To investigate human adaptation to a miniature robot during collaborative navigation tasks, exploring how both humans and robots can adapt to each other's behaviors. | Navigation task, gesture based | Two-way | • Performance | 22 |
| Stefan K Ehrlich and Gordon Cheng [33], 2018, Germany | It explores the usability of error-related potentials (ErrPs) as a feedback signal for mediating co-adaptation in less constrained human-robot interaction scenarios. | Guessing game, BCI | Two-way | • Performance | 18 |
| Dimova-Edeleva et al [34], 2022 | To investigate whether self-related and agent-related errors evoke different error-related potentials (ErrPs) in the human brain during human-agent collaboration. | BCI research, grid-world game | Two-way | • Shared workspace<br>• Responsibility | 11 |
| Lee Chang-Shing et al [35], 2022, Tiawan, Canada, Japan | It proposes an Adaptive Fuzzy Neural Agent (AFNA) with a Patch Learning Mechanism for human and machine co-learning using Fuzzy Markup Language (FML). | Go Game | Two-way | • Performance | 19 |
| Hu Xuhui et al [36], 2023 | To develop a human-robot co-adaptation framework | Automatic Control in | Two-way | • Performance<br>• Intent | 12 |

| Study, year, place of Publication | Objective | Application | | Cognitive theories and frameworks | Sample size, n |
|---|---|---|---|---|---|
| | using biofeedback-based user adaptive behavior to enhance user intent recognition in automatic control systems for assistance robots. | Assistive devices, wrist movement based on surface electromyogram (sEMG) | | recognition<br>• Robust control | |
| Lee Chang-Shing et al [37], 2020 | To introduce and evaluate a novel fuzzy-based system for human-robot cooperative Edutainment, integrating a brain-computer interface (BCI) ontology model and a Fuzzy Markup Language (FML)-based Reinforcement Learning Agent (FRL-Agent). | Human-Robot Cooperative Edutainment, Ontology | Two-way | • Feelings<br>• Attention<br>• Fatigue<br>• Stress | 10 |
| Karel van den Bosch et al [38], 2019, Netherlands | Discusses the challenges of enabling effective human-AI co-learning and collaboration in hybrid teams, emphasizing the need for bidirectional learning between humans and AI systems. | Challenges for Human-AI co-learning | Two-way | • Mental model<br>• Theory of minds<br>• Explainable AI<br>• Decision making | NA |
| van Zoelen et al [39], 2021, Netherlands | To identify interaction patterns of mutual adaptation in human-robot teams to facilitate co-learning, where both humans and robots learn from each other | Search and rescue (in earthquake operation) | Two-way | • Mental model<br>• Situation Awareness | 24 |
| Li Huao et al [40], 2021, United States | To develop an adaptive agent that can identify and complement human policies in human-agent teams to optimize team performance | Team Space Fortress | Two-way | • Decision making<br>• Trust | 104 |
| van Zoelen et al [41], 2023, Netherlands | To develop an ontology-based model and communication interface to enable human-AI teams to recognize, formalize, and communicate emergent collaboration patterns for | Human-agent team Ontology of Collaboration Patterns | Two-way | • Decision making<br>• Performance | 10 |

| Study, year, place of Publication | Objective | Application | | Cognitive theories and frameworks | Sample size, n |
|---|---|---|---|---|---|
| | effective co-learning | | | | |
| Ahlberg Sofie et al [42], 2022, Sweden | To investigate and summarize the challenges and methodologies for achieving co-adaptive human-robot cooperation, emphasizing the need for robots to adapt to humans in real-time. | Concept/ Summary: COIN project, safe planning/control | Two-way | • Trust,<br>• Mental model<br>• Decision making | NA |
| Shafti Ali et al [43], 2020, United Kingdom | To explore how humans and robots interact implicitly on a motor adaptation level in a collaborative maze game using deep reinforcement learning. | Motor adaptation level, Maze game | Two-way | • Performance<br>• Behavior | 7 |
| Azevedo Carlos R. B. et al [44], 2017, Brazil | To establish human-machine mutual understanding (HMMU) to foster trust and collaboration between humans and autonomous systems | Human-machine mutual understanding (HMMU), Autonomous Systems | Two-way | • Situation Awareness<br>• Cognitive Agents<br>• Trust | NA |
| Krinkin Kirill et al [45], 2023, Russia | It proposes a co-evolutionary hybrid intelligence model that integrates human and machine cognitive functions to overcome the limitations of data-centric AI | Summary: Cognitive architectures | Two-way | • Problem solving<br>• Mental structures<br>• Performances | NA |
| Yao Yiyu [46], 2023, Canada | To introduce the concept of the SMV (Symbols-Meaning-Value) space for understanding human-machine symbiosis and co-intelligence | Concept/ Summary: SMV (Symbols-Meaning-Value) | Two-way | • Decision making<br>• Mental model<br>• Trust | NA |
| Ansari Fazel et al [47], 2018, Austria | To explore mutual (reciprocal) learning to develop collective human-machine intelligence in smart factories | Concept/ Summary: ATODIDACT: Smart factory Industry 4.0 | Two-way | • Decision making<br>• Problem solving | NA |
| Liu Yizhi and Jebelli Houtan | It explores the feasibility of using an adaptive neural | Construction industry, | Two-way | • Performance<br>• Mental images | 4 |

| Study, year, place of Publication | Objective | Application | | Cognitive theories and frameworks | Sample size, n |
|---|---|---|---|---|---|
| [49], 2022, United States | network to translate workers' mental images into robotic commands via brainwave signals for controlling construction robots. | Brainwave signals- EEG | | | |
| Izadi Vahid et al [50], 2023, United States | To establish a framework for studying co-adaptation between humans and automation systems in a haptic shared control framework, focusing on resolving conflicts during co-steering of a semi-automated ground vehicle. | Automation system, Haptic shared control | Two-way | • Performance<br>• Resolving conflict | NA |
| Tsitos Athanasios C [51], 2022 | To enhance mutual performance in human-robot collaboration (HRC) settings by using a transfer learning technique called Probabilistic Policy Reuse, allowing a deep reinforcement learning (DRL) agent to adapt to different human partners in real-time. | Human-Robot Collaborative Game | Two-way | • Team performance<br>• Decision making | 16 |
| Xie Baijun and Park Chung Hyuk [53], 2023 | To develop a robotic intelligence framework using mutual learning paradigms, focusing on emotion recognition and behavior perception for enhancing social interactions in healthcare settings. | Social robotics in clinical and healthcare, game of charades, gesture recognition | Two-way | • Empathy(Social boding, stress, anxiety)<br>• Engagement | 5 |
| Yorita Akihiro et al [54], 2013 | To enhance the effectiveness of robot-assisted language learning by addressing issues of boredom through mutual learning based on social cognitive theory, focusing on adapting the robot's behavior to maintain engagement over long-term interactions. | Language Education, Robot-Assisted Language Learning | Two-way | • Communication<br>• Self-efficacy<br>• Social cognitive theory | 1 |

| Study, year, place of Publication | Objective | Application | | Cognitive theories and frameworks | Sample size, n |
|---|---|---|---|---|---|
| Ahtinen Aino et al [55], 2023 | To explore and evaluate co-learning activities between elementary school pupils and university students using social robots, focusing on educational benefits and data privacy considerations. | Social Robotics, Educational Robotics | NA | • Perspectives<br>• Learning | 56 |
| van Zoelen et al [56], 2021, Netherlands | To study human-robot co-learning processes to facilitate fluent collaborations through implicit adaptation and task learning. | Experiment 1- physical navigation task with the robot on a leash, Experiment 2- Search and rescue in virtual task | One-way and Two-way both | • Performance<br>• Mental model | NA |
| Ikemoto Shuhei et al [57], 2012, Japan, Germany | To present and evaluate a computationally efficient machine learning algorithm for improving interaction quality between robots and human caregivers in close-contact scenarios. | Assistive Tasks: human parenting behavior, namely, an assisted standing-up task and an assisted walking task | Two-way | • Judgement<br>• Dependable<br>• Safety<br>• Performance<br>• Efficiency | 5 |
| Dai Wei et al [58], 2021 | To investigate perspective-taking in human-robot interaction, focusing on how changes in the robot's cognitive-affective state influence children's behavior, emotional responses, and perception of the robot during co-learning interactions. | Brain-computer interface (BCI), EEG artifact | Two-way | • Efficiency<br>• Decision making<br>• Performance | 6 |
| Lee Hee Rin et al [59], 2017 | To develop a methodology for the participatory design of social robots that integrates feedback from older adults diagnosed with depression and their therapists, aiming for socially robust and | participatory design of social robots | Two-way | • Depression | 15 |

| Study, year, place of Publication | Objective | Application | | Cognitive theories and frameworks | Sample size, n |
|---|---|---|---|---|---|
| | responsible robot design. | | | | |
| Moreno J. C. et al [60], 2014 | To enable positive co-adaptation and more seamless interaction between humans and wearable robots (WRs), focusing on improving adaptability and flexibility through cognitive architectures. | Wearable robots, Bioinspiration, Neurorehabilitation | Two-way | • Symbiotic-gait behavior | NA |
| Peruzzini Margherita et al [61], 2023, Italy | To propose a Smart Manufacturing Systems Design (SMSD) framework enabling Industry 5.0 based on human-automation symbiosis. | Review: Smart Manufacturing Systems Design (SMSD) for Industry 5.0 | Two-way | • Decision making<br>• Socio-technical challenges<br>• Productivity | NA |
| Hodossy Balint K and Farina Dario [63], 2023 | To develop a Deep Reinforcement Learning-based motion controller for virtual testing of wearable robotic devices, focusing on co-adaptation and user intent representation in simulated prosthetic leg control. | Rehabilitation robotics | Two-way | • Robot-assisted gait<br>• Performance<br>• Decision making | 0 |
| Nikolaidis et al [64], 2017, Singapore | To evaluate how mutual adaptation between humans and robots affects team performance and collaboration during shared autonomy tasks. | Bounded-Memory Adaptation- an assistive robotic arm on a table-clearing task | Two-way | • Performance<br>• Trust<br>• Stochastic decision model | 51 |
| Yadollahi Elmira et al [65], 2019 | To propose and evaluate a shared control framework for human-multirobot foraging using a brain-computer interface (BCI), aiming to integrate human operator opinions with robot consensus to improve cooperation efficiency in multi-robot systems. | Game of goal cards | Two-way | • Perspective taking<br>• Interactions, Engagement<br>• Emotional state<br>• Frustration | NA |
| Yamagami Momona [67], | To develop individualized adaptive algorithms using | Bio-signal, EMG, HMI | Two-way | • User-centered design- | 12 |

| Study, year, place of Publication | Objective | Application | | Cognitive theories and frameworks | Sample size, n |
|---|---|---|---|---|---|
| 2022 | multi-channel biosignals for improving accessibility and health outcomes through human-machine interfaces (HMIs). | | | controller | |
| Ren Yuqing et al [68], 2023, United States | To explore the complementarity between human and AI capabilities and identify optimal ways to structure and coordinate their collaboration | Human and AI complementarity | Two-way | • Metal knowledge<br>• Human skills | NA |
| Ogiso Takaya et al [69], 2015, Japan | It proposes a hybrid learning system that facilitates collaborative learning between humans and artificial intelligence systems to accelerate learning in unknown environments | Humans and artificial intelligence systems for color design task smart-grid game | Two-way | • Enhancing skills<br>• Accuracy | 20 |
| Min Hun Lee et al [70], 2021, United States | To develop and evaluate an interactive AI-based system that collaborates with therapists for more accurate rehabilitation assessments | Healthcare information systems, task oriented upper limb exercises | Two-way | • Decision making<br>• Decision support system<br>• Mental model<br>• Trust<br>• Attention | 26 |
| Woolley Anita Williams et al [71], 2023, United States | To understand how technology-enabled, distributed, and dynamic collectives can be designed to solve complex problems over time, focusing on the Transactive Systems Model of Collective Intelligence (TSM-CI). | Phycology-Transactive Systems Model of Collective Intelligence (TSM-CI), emergence and maintenance | Two-way | • Attention<br>• Reasoning<br>• Memory | NA |
| de Boer Maaike H.T. et al [72], 2022, Netherland | To present the FATE system for decision support using human-AI co-learning, explainable AI, and privacy-preserving data usage in various real-world applications. | FAIT system: Fair, Transparent and Explainable Decision Making in a Juridical Case | Two-way | • Explainable AI<br>• Decision making<br>• Trust | 200 |
| Kirill Krinkin et al | Discusses an alternative | Concept/ | Two-way | • Decision | NA |

| Study, year, place of Publication | Objective | Application | | Cognitive theories and frameworks | Sample size, n |
|---|---|---|---|---|---|
| [73], 2023, Russia | approach to developing artificial intelligence systems through human-machine hybridization and their co-evolution. | Summary: Applied medicine, MIS | | making<br>• Cognition | |
| Alessia Vignolo et al [74], 2019, Italy, Scottland | To investigate the impact of a robot teacher's effortful adaptation on human learning and rapport in a skill-teaching scenario. | YARP (Yet Another Robot Platform) | Two-way | • Effort<br>• Effectiveness<br>• Rapport<br>• Performance<br>• Perception | 21 |
| Gallina Paolo et al [75], 2015 | To explore the concept of co-adaptation between a human operator and a machine interface, emphasizing its application in teleoperation and assistive technology, with a focus on incorporating temporal evolution of co-adaptation parameters. | Human centered design for interface, Telemanipulati on, Active Vision, | Two-way | • Usability<br>• Performance | NA |
| Winkle Katie et al [76], 2020 | To present a mutual shaping approach to the design of socially assistive robots, focusing on their use in therapy and emphasizing mutual learning and societal factors. | social robots in therapy, user-centered and participatory design | Two-way | • Sharing and shaping of knowledge<br>• Ideas<br>• Acceptance | 19 |
| Ehrlich Stefan K. et al [77], 2019 | To develop a computational model to study the dyadic interaction between humans and robots using error-related potentials (ErrPs), focusing on how variations in ErrP-decoder performance affect co-adaptation in human-robot interaction (HRI). | Neuroscience | Two-way | • Decision making<br>• Performance<br>• Perception<br>• Learning rates | 16 |
| Markelius Alva et al [78], 2024 | To establish a human-robot interaction setup where both entities learn a symbolic language for identifying robot homeostatic needs, using a | homeostatic motivational grounding of the robot’s language, Socially | Two-way | • Empathy<br>• Affective experience | 6 |

| Study, year, place of Publication | Objective | Application | | Cognitive theories and frameworks | Sample size, n |
|---|---|---|---|---|---|
| | differential outcomes training (DOT) protocol. | Assistive robots | | | |
| Chang-Shing Lee, et al [79], 2020, Tiawan, Japan | To develop an intelligent agent for robotic edutainment and humanized co-learning, integrating brain-computer interface technology with AI bots for real-world applications. | Edutainment, Go players, OpenGo darkforest, Brainwaves | Two-way | • Learning Performance<br>• Attention | 5+classroom (NA) |
| Jokinen Kristiina and Watanabe Kentaro [80], 2019 | To explore the concept of Boundary-Crossing Robots, focusing on their interaction capabilities and their role in symbiotic interactions with human users in everyday activities. | Boundary-crossing robot, symbiotic interactions | Two-way | • Trust | NA |
| Schoonderwoerd et al [81], 2022, Netherlands | To develop and evaluate Learning Design Patterns (LDPs) to facilitate human-AI co-learning in a simulated urban-search-and-rescue task. | Ontology-based AI-model | Two-way | • Decision making<br>• Trust<br>• Mental model | 35 |
| Petric Tadej et al [82], 2017 | To explore how two individuals adapt their motor behavior while physically interacting through an object, and to propose a novel control algorithm for robots in human-robot cooperative setups based on these findings. | physically interactive manipulation task | Two-way | • Human-robot interaction<br>• Motor behavior<br>• Performance | 20 |

Table 1 shows that about 89.61% (69/77) of the studies have focused on promoting co-learning and mutual adaptation in recent years. Additionally, many of these studies have observed cognitive behaviors such as decision-making, trust, mental models, and improved performances (76.62%) when completing tasks as part of a human-agent team. For example, the collaboration between humans and AI has helped improve healthcare information systems by developing a decision support system, thus enhancing trust and decision-making and analyzing different mental models for patients' upper limb exercises [70]. There has also been improved efficiency and decision-making through a shared control framework in human-multi-robot foraging collaboration with the Brain-computer interface (BCI) [58], providing a more comprehensive perception during co-learning interactions. Table 2 provides the intelligent agent/artificial intelligence (AI) methods used in the selected studies for co-learning/co-adaptation, along with performance measures if reported (with "NA" indicating not available).

**Table 2.** Intelligent Agent/Artificial intelligence (AI) methods used in and performance measures for co-learning/co-adaptation.

| **Study, year** | **Intelligent Agent/AI method or Algorithm** | **Team Performance Metric** |
|---|---|---|
| Emma M. et al [1], 2021 | • Remotely controlled robot with leash | NA |
| Stefanos Nikolaidis et al [2], 2017 | • Mixed Observability Markov Decision Process (MOMDP) | • Task Completion<br>• Satisfaction |
| Xu Yong et al [3], 2009 | • WAITER (Waiter Agent Interactive Training Experimental Restaurant) - software simulation system<br>• Instruction based system | • Effectiveness |
| Xu Yong et al [4], 2011 | • WAITER | • Effectiveness |
| Xu Yong et al [5], 2009 | • WAITER | • Response behavior |
| Yong Xu et al [6], 2012 | • WAITER- Matlab based computer game<br>• Bayesian network model | NA |
| Ramya Ramakrishnan et al [7], 2017 | • Transfer Learning<br>• RL<br>• Policy Reuse in Q-Learning (PRQL) algorithm | NA |
| Döppner et al [8], 2019 | • The artifact for unit load device in Air Cargo Industry | NA |
| Burke C. S. et al [9], 2006 | • NA | NA |
| Hendrik Buschmeier and Stefan Kopp [10], 2013 | • Bayesian Network | • Effectiveness |
| Guy Hoffman [11], 2019 | • Robot/Agent- Markov Decision Process (MDP) | • Human-Robot fluency |
| Lygerakis Fotios et al [12], 2021 | • Soft Actor-Critic agent<br>• RL framework | • Accuracy<br>• Efficiency<br>• Effectiveness<br>• Total Training Time |
| Zhao Michelle et al [13], 2022 | • POMDP | • Effectiveness<br>• Task completion efficiency<br>• Subjective evaluation of the human-AI interaction quality |
| Nikolaidis et al [14], 2016 | • MOMDP | • Effectiveness |
| Youssef Khaoula et al [15], 2014 | • Dish Robot<br>• Actor-critic architecture | • Effectiveness of communication protocols established between the human and the dish robot<br>• Adaptation accuracy to individual preferences<br>• Overall efficiency and satisfaction |
| Damian Dana D. et al [16], 2009 | • Simulated Robotic Hand | • Effectiveness |
| Zhou Hua et al [17], 2021 | • UR5 Robot<br>• Virtual Fixtures | • Cognitive load<br>• Satisfaction |
| Ogata T et al [18], 2005 | • Humanoid Robot- Robovie<br>• FFNN<br>• RNN for robot control | • Effectiveness, efficiency, accuracy |
| De Santis Dalia [19], 2021 | • Unsupervised learning | • Efficiency |
| Yamada S and Yamaguchi T [20], 2002 | • Microsoft agent<br>• Nearest Neighbor (NN) method,<br>• SONY pet robot AIBO | NA |
| Madduri Maneeshika M. et al [21], 2022 | • Decoder adaptation used a supervised learning algorithm. | • Learning rates<br>• Effectiveness |
| Sciutti Alessandra and Sandini Giulio [22], 2017 | • Humanoid Robot-iCub | • Efficiency |
| Gao Yuan [23], 2020 | • Deep RL | • Learning process |
| Young Carol et al [24], 2019 | • Dual Expert Algorithm (DEA)<br>• Human Aware Dual Expert Algorithm (HADEA) | • Stability<br>• Accuracy |
| O'Meara Sarah M. et al [25], 2021 | • K Nearest Neighbors (KNN) | • Accuracy |
| Warrier Rahul B. and Devasia Santosh [26], 2017 | • Robot-arm | • Tracking error<br>• Output tracking |
| Damiano Luisa and Dumouchel Paul [27], 2018 | NA | NA |
| | | |
| Cress Ulrike and Kimmerle Joachim, [28], 2023 | Generative AI | NA |
| Ahtinen Aino et al [29], 2023 | • Alpha mini robot | • Effectiveness |
| Kitagawa Masahiro et al [30], 2016 | • Robovie-R ver3<br>• Humanoid Robots- Pepper | • Effectiveness |
| Mohammad Yasser and Nishida Toyoaki [32], 2008 | • E-puck designed by EPFL | • Effectiveness<br>• Feedback |
| Stefan K Ehrlich and Gordon Cheng [33], 2018 | • NOA Robot<br>• ErrP | • Effectiveness<br>• Efficacy<br>• Efficiency |
| Dimova-Edeleva et al [34], 2022 | • Support Vector Machines (SVM) | • Accuracy |
| Lee Chang-Shing et al [35], 2022 | • Adaptive Fuzzy Neural Agent (AFNA) with a Patch Learning Mechanism<br>• Fuzzy Markup Language (FML) | • Regression Model |
| Hu Xuhui et al [36], 2023 | • Biofeedback in an interactive interface<br>• Intent recognition model | • Completion time<br>• Efficiency<br>• Robustness |
| Lee Chang-Shing et al [37], 2020 | • Fuzzy Markup Language(FML)-based RL Agent with Fuzzy Ontology | • Learning progress |
| Karel van den Bosch et al [38], 2019 | NA | NA |
| van Zoelen et al [39], 2021 | • Reinforcement Learning (RL) Virtual ROBOT | NA |
| van Zoelen et al [41], 2023 | • Drag-and-Drop Graphical User Interface (GUI) | • Collaboration Pattern<br>• Usability |
| Ahlberg Sofie et al [42], 2022 | • RL<br>• LTL | NA |
| Shafti Ali et al [43], 2020 | • Universal Robots UR10- robotic manipulator<br>• RL | • Success Rate of solving the collaborative maze game |
| Azevedo Carlos R. B. et al [44], 2017 | • Autonomous systems | NA |
| Krinkin Kirill et al [45], 2023 | • AI | NA |
| Yao Yiyu [46], 2023 | • Summary of recent works | NA |
| Ansari Fazel et al [47], 2018 | • AUTODIDACT<br>• Machine Digital Twin | • Learning outcomes<br>• Learning curves |
| Liu Yizhi and Jebelli Houtan [49], 2022 | • Masonry robot<br>• Adaptive Neural Network | • Success rate<br>• Effectiveness |
| Izadi Vahid et al [50], 2023 | • RL-based Model Predictive Controller (MPC) | • Effectiveness<br>• Task completion efficiency<br>• Accuracy of steering<br>• Subjective evaluations of collaboration quality |
| Tsitos Athanasios C [51], 2022 | • ROS<br>• Deep RL-Soft actor-critic algorithm<br>• Universal Robot UR3 | • Effectiveness |
| Xie Baijun and Park Chung Hyuk [53], 2023 | • Deep Learning based on emotion recognition<br>• Humanoid Robot-Pepper<br>• Convolution Neural Network (CNN) | NA |
| Yorita Akihiro et al [54], 2013 | • Fuzzy control<br>• MOBiMac<br>• Steady-State Genetic Algorithm (SSGA)<br>• Spiking Neural Networks (SNN)<br>• Self-Organizing Map (SOM)<br>• RL | • Effectiveness |
| Ahtinen Aino et al [55], 2023 | • Nao robot | • Effectiveness |
| van Zoelen et al [56], 2021 | • Experiment 1- Robot on a leash<br>• Experiment 2- RL | • Effectiveness<br>• Learning Outcomes<br>• Interactive Patterns |
| Ikemoto Shuhei et al [57], 2012 | • Child–robot with biomimetic body<br>• ML-Gaussian mixture models (GMMs) | • Task completion accuracy<br>• Efficiency<br>• Subjective perception of the interaction quality |
| Dai Wei et al [58], 2021 | • Multi-Robot Systems(MRS)<br>• Steady-State Visually Evoked Potentials (SSVEP) | • Efficiency |
| Lee Hee Rin et al [59], 2017 | • Socially Assistive Robots (SARs) | • Engagement<br>• Effectiveness |
| Moreno J. C. et al [60], 2014 | • Neuro-Musculo-Skeletal (NMS) simulator | NA |
| Peruzzini Margherita et al [61], 2023 | • SMSD framework | • Factory Productivity<br>• Workers well being |
| Hodossy Balint K and Farina Dario [63], 2023 | • DRL based motion controller MuJoCo | • Tripping rate<br>• Gait patterns |
| Nikolaidis et al [64], 2017 | • Mixed Observability Markov Decision Process (MOMDP) | • Mean reward for the trials<br>• Kruskal-Wallis H test |
| Yadollahi Elmira et al [65], 2019 | • NAO robot | • Efficiency |
| Yamagami Momona [67], 2022 | • Game theory to model | NA |
| Ren Yuqing et al [68], 2023 | • Summary of recent works | NA |
| Ogiso Takaya et al [69], 2015 | • General Regression Neural Network (GRNN) | • Accuracy using Euclidean distance |
| Woolley Anita Williams et al [71], 2023 | • Transactive Systems Model of Collective Intelligence (TSM-CI) framework (review) | NA |
| de Boer Maaike H.T. et al [72], 2022 | • FAIT system | NA |
| Kirill Krinkin et al [73], 2023 | • AI based Intelligent systems considered example of applied medicine to describe the stages of CHI<br>• Summary of the concept | NA |
| Alessia Vignolo et al [74], 2019 | • Humanoid robot iCub, | • Effectiveness of Human Teacher<br>• Perception of Robot Teacher<br>• Rapport of Both |
| Gallina Paolo et al [75], 2015 | • User interface | • Task completion rate<br>• Error rates<br>• Efficiency of interaction<br>• User Feedback on usability and adaptability |
| Winkle Katie et al [76], 2020 | • SARs | • Effectiveness<br>• Learning process |
| Ehrlich Stefan K. et al [77], 2019 | • RL- Actor critic architecture | • Effectiveness<br>• Efficacy<br>• Efficiency |
| Markelius Alva et al [78], 2024 | • Differential outcomes training<br>• SDK-version of the Reachy robot | • Learning efficiency |
| Chang-Shing Lee, et al [79], 2020 | • Robot Palro with an intelligent agent | • Learning Performances<br>• Feedback |
| Jokinen Kristiina and Watanabe Kentaro [80], 2019 | • Social Robots | NA |
| Schoonderwoerd et al [81], 2022 | • Learning Design Patterns (LDPs) | NA |
| Petric Tadej et al [82], 2017 | • ML-Dynamic Movement Primitives (DMPs) | • Efficiency<br>• Accuracy |
| Li Huao et al 2021, [40] | • Partially Observable Markov Decision Process (POMDP) | NA |
| Min Hun Lee et al 2021, [70] | • Markov Decision Process- Q-network with Double Q-learning<br>• Machine Learning (ML)<br>• Rule based | • Precision<br>• Recall |

* NA- Not Available

### Human-Agent/AI Teams

The concept of Human-Agent/Robot/AI has been there for over a decade in the field of Human-Robot Collaboration. However, the research has primarily focused on one-sided adaptation, either by the robot or the human, such as in medical robotics and assistive technology [62], [66]. Thus, to answer RQ1, our review explicitly focuses on two-sided adaption, where team members can learn from each other and adapt to any task. For this reason, we have depicted the two adaptation styles in Figure 3 in this review, showcasing the major adaptation characteristics and the commonly analyzed cognitive frameworks observed. We found that, among 82 papers reviewed, 69 (84.14%) focused on the two-way adaptation style, 5 (6.09%) on one-way, 4 (4.87%) described both, and 4 (4.87%) did not report any style explicitly. The first-ever study on mutual adaptation was published in 2002, in which a Microsoft agent and a human played a game of mutual mind reading, where they could predict each other's mind state on a GUI [66]. Since then, research in this field has been limited, with an average of one paper per year until 2017. These mutual adaption applications mostly revolved around WAITER robots [3-5], boundary memory adaptation, rehabilitation robotics, and language education [14], [16], [54]. The term co-learning was first introduced by authors of [69] in 2015, proposing a hybrid learning system for human AI to accelerate learning in an unknown environment through a color-designed smart-grid game. It wasn't until 2021 that the concept of Human-Agent/Robot/AI collaboration gained significant traction and piqued the interest of researchers. This shift was further propelled by Zoelen et al., who were the first to define and popularize co-learning, co-adaptation, and co-evolution properly. They did so by drawing inspiration from a real-life earthquake scenario and virtually designing an experiment for a search and rescue operation, where humans and robots work together to complete a task [39].

With the recent surge in AI and robotics, we have witnessed a significant upsurge in the trend of Human-Agent teaming to co-learn and co-adapt. Figure 4 provides a snapshot of the highest number of contributions in 2023. Notably, the foundation of this research area was laid back in 2002 by Yamada and Yamaguchi [20]. However, it took twenty significant years to see the growth of 64.93% (50/77) in the last five years, marking a significant leap in the field.

Moreover, the applications of Human-agent teams were observed to be widespread throughout various locations. What's particularly significant is the interest in studying human-agent teaming systems, both academically and in the industries. Primary authors and universities from the United States (18/77, 23.37%), Japan (14/77, 18.18% ), Netherlands, and Germany (7/77, 9.09%) each, United Kingdom (5/77, 6.49%), Italy and China (3/77, 3.89%) each, Finland, Russia, and Sweden (2/77, 2.59%), Austria, Brazil, Canada, Greece, Portugal, Singapore, Slovenia, Switzerland, (1/77, 1.29%) each, have all contributed to this interest. These countries have also worked collaboratively (11/77, 14.28%), further emphasizing the global significance and impact of the research.

**Figure 3. Comparing different adaptation styles for Human-Agent Teams.**

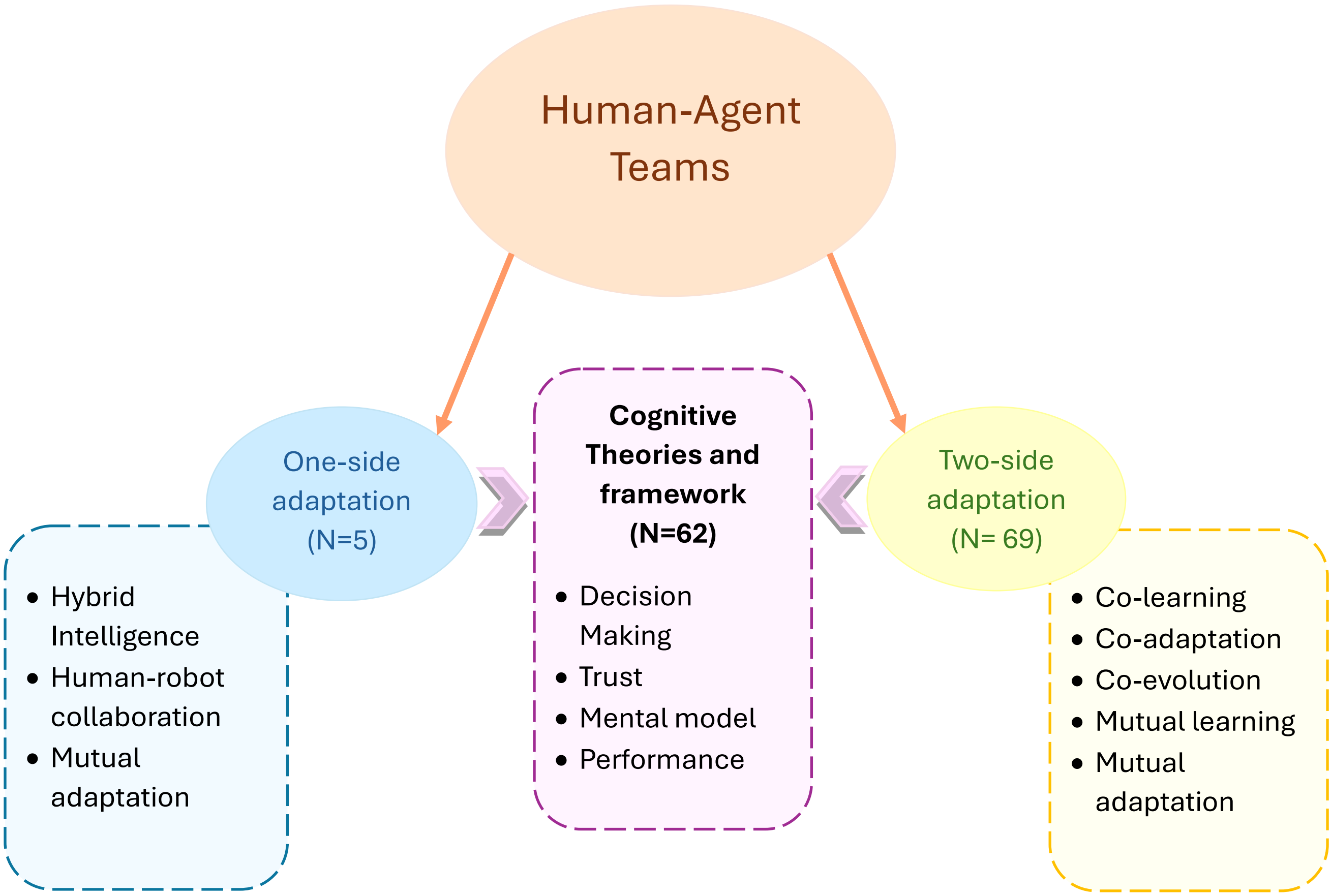

**Figure 4. The rising trend of co-learning and co-adaptation in Human-Agent Teams.**

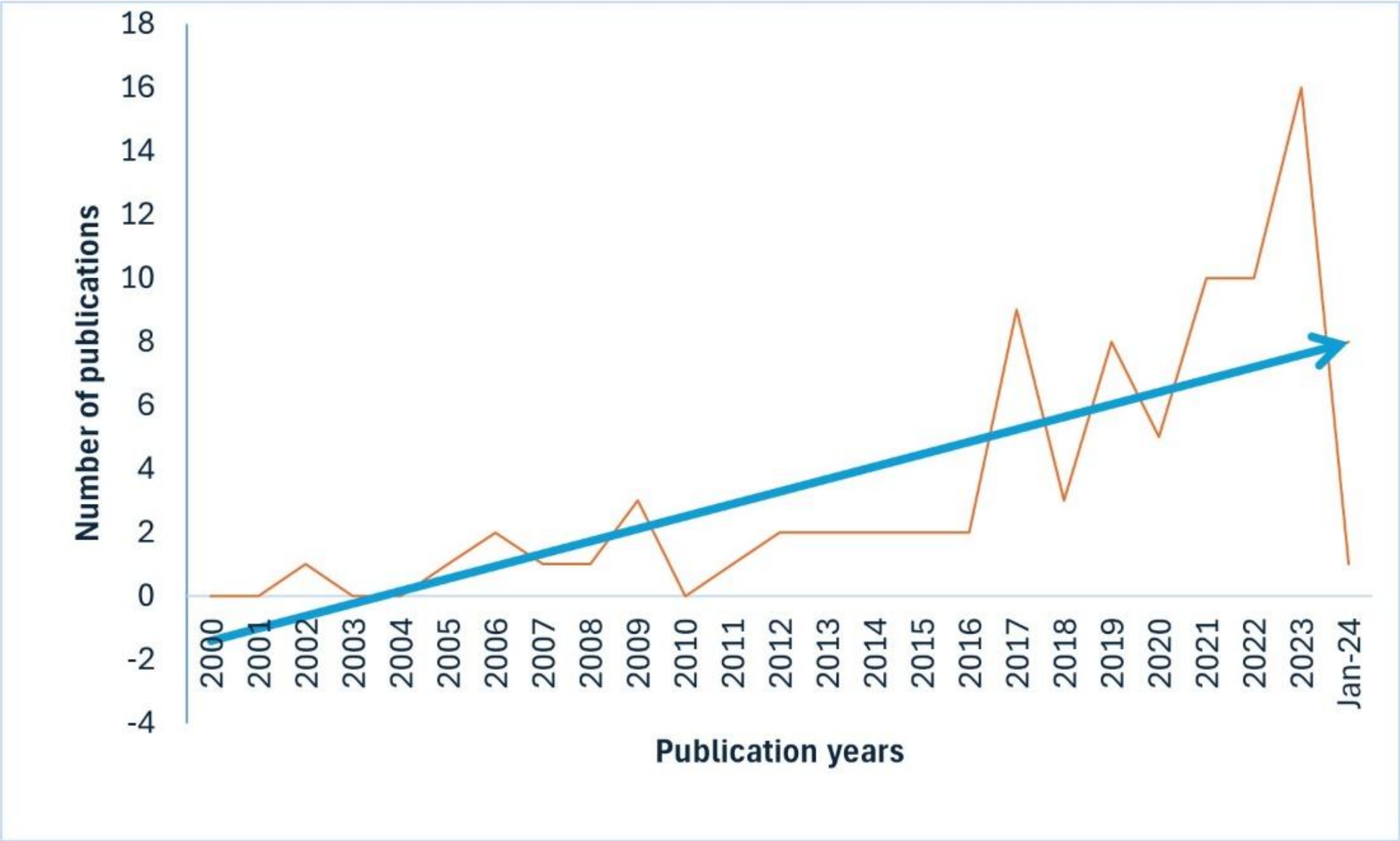

## Cognitive frameworks used by Human-Agent team members

In the 77 studies reviewed on human-agent co-learning and co-adaptation systems that reported participants' data, a wide array of cognitive theories and frameworks were explored. Most studies (61/77, 79.22%) focused on cognitive theories related to decision-making, trust, performance, and mental state. Other cognitive theories investigated include socio-cognitive engineering, perspective and learning, empathy and feelings, shared space and knowledge, completion time and learning experience, and communication. This diverse range of cognitive theories adds depth and richness to the research. Some studies quantitatively measured attitudes using instruments such as the Negative Attitude toward Robots Scale (NARS) [53], NASA TLX [18], or the 5-point Likert scale [6], [13], [14], [31], [64], [76], [81]. In contrast, a few studies (10/77, 12.98%) collected data through qualitative methods such as interviews, verbal feedback, think-aloud methods, and video recordings [1], [10], [39], [40].

Additionally, in this review, collaborative approaches were found to have a positive impact on various cognitive behaviors related to task completion. Human agents demonstrated improvements in task completion satisfaction [7], factory productivity and work well-being [61], success rate of task completion [43], [75], and learning rates and processes [21], [24], [37]. Additionally, there were reports of improved performance (32/77, 41.55%), decision-making (29/77, 37.66%), building better trust (18/77, 23.37%), and enhancing mental models (13/77, 16.88%). Furthermore, a few studies highlighted the significance of perspective and learning (3/77, 3.89%), empathy, feelings, shared space and knowledge, and socio-cognitive engineering (2/77 each) in co-learning and co-adaptation. For instance, social robots in therapy, designed with a user-centered and participatory approach, were observed to facilitate the effective sharing and shaping of knowledge, ideas, and acceptance [76]. Another study indicated the relevance of analyzing the impact of robots' frustration on children's perspectives, interactions, engagement, and emotional states to enhance co-learning [65]. The sample size of the studies can be found in Table 1.

The sample size ranged from 1 to 200 participants. Approximately half (15/77, 19.48%) of the collected studies had <10 participants and 29.87% of the studies (23/77) had no participants available. For example, Jokinen discussed applications and the societal impacts of boundary-crossing robot interactions with autonomous agents capable of social interactions to build trust [80]. The average sample size was 20.04 (SD 30.93) participants in the studies conducted from 2002 to 2019 and 31.59 (SD 43.10) participants from 2020 to 2024. This clearly indicates the increase in the trust and importance of successfully utilizing the human-agent teams for co-learning and co-adaptation for various critical applications.

## Concept and functions of Human-Agent/AI Teams

### Concept of mutual adaptation and co-learning

Human-agent teaming, a dynamic and evolving partnership, is a concept that heavily relies on mutual adaptation, co-adaptation, co-learning, mutual learning, and co-evolution. The studies in this review present systems based on this concept, with mutual adaptation (26/77, 33.76%), co-adaptation (19/77, 24.67%), co-learning (17/77, 22.07%), mutual learning (13/77, 16.88%), and co-evolution (9/77, 11.68%) being the key elements. This relationship not only emphasizes the functionality and efficiency of task completion but also the intricate dynamics of two-way learning and adaptation that emerge from continuous interaction. However, Yong describes the mutual adoption of gesture-based human-robot interfaces based on the Wizard of Oz experiment, which is a one-sided approach [19]. Similarly, in [48] and [66], the authors describe mutual learning and mutual adaptation of a hierarchical ad hoc agent (HAHA) that adapts to the human and an assistive device for restoration of motor function through EMG-based prosthetic hand that recognizes motion patterns, are both one-sided approaches, respectively.

### Types of Agents and AI Algorithms

In the selected studies, various agents in the form of physical robots and AI algorithms were used, as presented in Table 2. The studies considered humanoid robots such as NOA robot [33], Parlo [79], iCub [74], Dish robot [15], Pepper [53], Robovie [18], Sony pet robot ABIO [20], Socially Assistive Robots [59], Alpha mini robot [29], robotic arms such as UR5 robotic manipulator [17], KUKA LWR [31], as well as UAVs and UGVs [52] were physically deployed in these studies. The most widely used AI-based algorithms included Reinforcement Learning (21/77, 28.57%) and Bayesian Networks (4/77, 5.19%). It's worth noting that most of the studies (15/21) that implemented reinforcement learning were conducted recently due to the recent development in AI, explainable AI, and exceptional performance [38]. The wide range of applications of reinforcement learning, from navigation to healthcare, as shown in Table 2, is a testament to its potential. Nonetheless, reinforcement learning in these systems aims to accurately predict and assist humans with different tasks and initiate an atmosphere of mutual adaptation, learning, and understanding.

The second most used algorithm in the reviewed studies was Bayesian network algorithms. The studies using Bayesian networks focused on restaurant-based WAITER systems [2-5], with a primary focus on mutual decision-making, trust, and mental models. However, these studies consistently aimed to utilize the robot's previous behaviors or conditions to provide personalized suggestions, highlighting the challenge of limited adaptability in this algorithm.

Other algorithms used include learning design patterns [81], general regression neural networks [69], adaptive fuzzy neural agents [35], adaptive neural networks [49], Gaussian mixture models [57], deep learning [53], unsupervised learning [19], nearest neighbors [20], supervised learning [21], human-aware dual expert algorithm [24], support vector machine [34], k-nearest neighbors [25], and Generative AI [28], ML-dynamic movement primitives [82].

**Performance metrics**

It is essential to consider the performance metrics of both human and AI agents to understand how effectively they collaborate. Various performance metrics used in the studies are effectiveness, efficiency, accuracy, and learning rates. These studies reported effectiveness was 31.16% (24/77), efficiency was 19.48% (15/77), accuracy was 12.98% (10/77), and learning rates were 2.59% (2/77). These metrics indicate how well human-agent teams perform in a collaborative environment, where human and AI agents must adapt and learn from each other. Although these studies exploited various performance metrics in their research, they needed more clarity on the evaluating grounds. For instance, in [57], accuracy and efficiency were reported based on the subjective perception of the interaction quality. This disparity underscores the persistent challenge in human-AI collaboration: guaranteeing that human and agent/AI elements can function with great precision and dependability, particularly in crucial decision-making applications.

# *Discussion*

**Principal Findings**

Our scoping review, the first of its kind, delves into the intriguing realm of Human-Agent co-learning and co-adaptation. We explore the research questions outlined in the introduction, which delve into the existing collaboration patterns and interactions between human-agent team members, their adaptation style, and the cognitive theories and frameworks associated with their teaming behavior. Our findings offer crucial insights into the development of co-learning and co-adaptation of Human-Agent relationships, a field rapidly gaining traction among researchers. We also highlight the diverse applications of this research, from healthcare to assistive technologies, medical robotics, and search and rescue operations, all aimed at enhancing decision-making, trust, and performance in crucial tasks where humans and agents collaborate.

From this review, we could also infer the geographic locations of the publications were spread among Europe (39/77, 50.64%), Asia (26/77, 33.76%), North America (21/77, 27.27%), and South America (1/77, 1.29%). These are limited because most studies are lab-based experimental setups, and fewer have been tested in real-life applications. However, family-based co-learning with a social robot in home environments is helping children, older people, and adults to accomplish daily tasks at home [29]. Furthermore, only 10.38% (8/77 papers have reported of the studies have proved to have a significant sample size. In contrast, many studies included a small sample size of humans and agents adapting to each other. Thus, there is a pressing need for more conclusive evidence to validate the two-way learning and adapting approaches from the outcomes, highlighting the current gaps in the research and guiding future studies.

While most studies we identified focused on a two-way approach, the future holds even more promising opportunities. We emphasize the importance of conducting future studies to investigate the adoption and sustainable use of co-learning for humans and agents over an extended period. Such studies will not only improve performance but also enhance the trust and decision-making capabilities of the team. We have also noticed among the AI agents that reinforcement learning (RL) based agents are the top choice of the researchers, as most of its algorithms work on the principle of Q-learning and the Markov decision process. Nevertheless, advancements in AI have led to the involvement of deep reinforcement learning (DRL) agents and generative AI methods to learn in real-time and solve shared tasks through efficient collaboration, which are still in their exploratory phase.

Moreover, this review found that 12% of the study focused on accuracy. Still, it was unclear whether the accuracy of task completion, any specific parameter, or the algorithm's performance was being measured. Nonetheless, this limiting factor was challenging to report and investigate.

## Implications and Future Work

A team becomes skilled by adapting and learning together, referred to as co-learning. In human teams, partners naturally adjust to each other and learn while working together. However, this only sometimes happens in human-robot teams [56]. So, the need arises to describe and support co-learning in human-robot partnerships. To address this challenge, it is essential to implement a system that enhances human and machine learning to support various industries, including manufacturing, healthcare, robotics, and assistive technology. In this review, we have emphasized the importance of a two-way approach to learning and adaptation in the context of advancing AI technology. Teams equipped with this approach can enhance collaborative performance and contribute to developing a mental model of the human partner's learning process. This, in turn, can help the robot effectively learn and adapt to the human partner. Ultimately, this collaborative work dynamic can improve decision-making, trust, and overall performance within human-agent teams, underscoring the integral role of the proposed system in supporting and advancing various industries.

Our future work will be centered on developing learning profiles and trajectories for both human and machine workforce, with a specific focus on human-robot collaboration, maintenance, and assembly use cases as mentioned in TU Wien's Pilot Factory Industry 4.0 [47]. This will involve outlining system specifications for modeling and measuring mutual learning, establishing an ontological knowledge base to specify shared conceptualization of tasks and associated domain knowledge, and defining rules for optimal task sharing and measuring learning outcomes. Furthermore, the need encompasses research areas including XAI, co-learning, fairness and bias, and secure learning, demonstrating its general applicability, for instance, refining the FATE systems through various domain-specific use cases [72].

## Limitations

Due to our evaluation of publications up to January 2024, this review only covers publications before that date, which is a limitation of this study. We focused on the two-way adaptation styles and corresponding cognitive behaviors between humans and agents. As a result, some applications involving one-sided human-robot interactions were not included in the review. In future studies, it's essential to prioritize research areas such as explainable AI, fairness and bias, and secure learning while also considering co-learning and co-adaptation. For instance, the FATE system is a general decision support system that addresses bias and fair AI, explainable AI, co-learning, and secure learning in domains like diabetes and the judiciary [72].

## Conclusions

Our review found studies on human-agent co-learning and co-adaptation, exploring different definitions of "co-" and "mutual" learning and adaptation. These studies reflect two styles of adaptation and cognitive theories applied across various fields. We compared two perspectives on co-adaptation and highlighted the need for more research. With the advancement of AI and robotics, human-machine interactions have significantly expanded since 2020. Studies have shown that human-agent teamwork is increasing worldwide and has the potential to solve real-life problems more efficiently while increasing trust among team members. However, many studies had small sample sizes. In the future, research should focus on identifying challenges, including the black box issue of AI and sustainable use of this approach, and seek solutions identified in related fields.

## Conflicts of Interest

None declared.

## Multimedia Appendix 1

PRISMA (Preferred Reporting Items for Systematic Reviews and Meta-Analyses) guidelines.

PRISMA_DOCX_file.docx

[DOCX File , 31 KB-Multimedia Appendix 1]